\documentclass{article} 
\usepackage{iclr2027_conference,times}

\usepackage{amsmath,amsfonts,bm}

\def\eqref#1{equation~\ref{#1}}

\def\1{\bm{1}}

\DeclareMathAlphabet{\mathsfit}{\encodingdefault}{\sfdefault}{m}{sl}
\SetMathAlphabet{\mathsfit}{bold}{\encodingdefault}{\sfdefault}{bx}{n}

\usepackage{hyperref}
\usepackage{url}

\usepackage{graphicx}
\usepackage{booktabs}
\usepackage{multirow}
\usepackage[table,dvipsnames]{xcolor}
\usepackage{amssymb}
\usepackage{amsmath}
\usepackage{arydshln}
\usepackage{xcolor}
\usepackage{wrapfig}

\definecolor{bestcolor}{RGB}{220,235,255}
\definecolor{secondcolor}{RGB}{235,235,235}

\newcommand{\methodname}{EgoHSI3R}

\title{\methodname{}:  Online 4D Hand-Scene Interaction Reconstruction from Egocentric Video}

\author{%
\makebox[\dimexpr\textwidth-2\tabcolsep\relax][c]{%
\normalfont\normalsize\bfseries
Mingzhi Pei\textsuperscript{*}\enspace \enspace
Zhangchen Ye\textsuperscript{*}\enspace \enspace
Ruoqu Chen\enspace \enspace
Liu Cao\enspace \enspace
Mengdi Xu\textsuperscript{$\dagger$}}\\[0.5em]
\makebox[\dimexpr\textwidth-2\tabcolsep\relax][c]{\normalfont
Tsinghua University}
\\[0.5em]
\makebox[\dimexpr\textwidth-2\tabcolsep\relax][c]{\normalfont
* Equal contribution\qquad $\dagger$ Corresponding author}
\\[1em]
\makebox[\dimexpr\textwidth-2\tabcolsep\relax][c]{%
    \normalfont\normalsize
    \href{https://pmz-enterprise.github.io/egohsi3r-page/}{%
        \nolinkurl{https://pmz-enterprise.github.io/egohsi3r-page/}%
    }%
}
}

\iclrpreprintcopy
\begin{document}

\maketitle

\begin{figure*}[htbp]
    \vspace{-9mm}
    \centering
    \includegraphics[width=\textwidth]{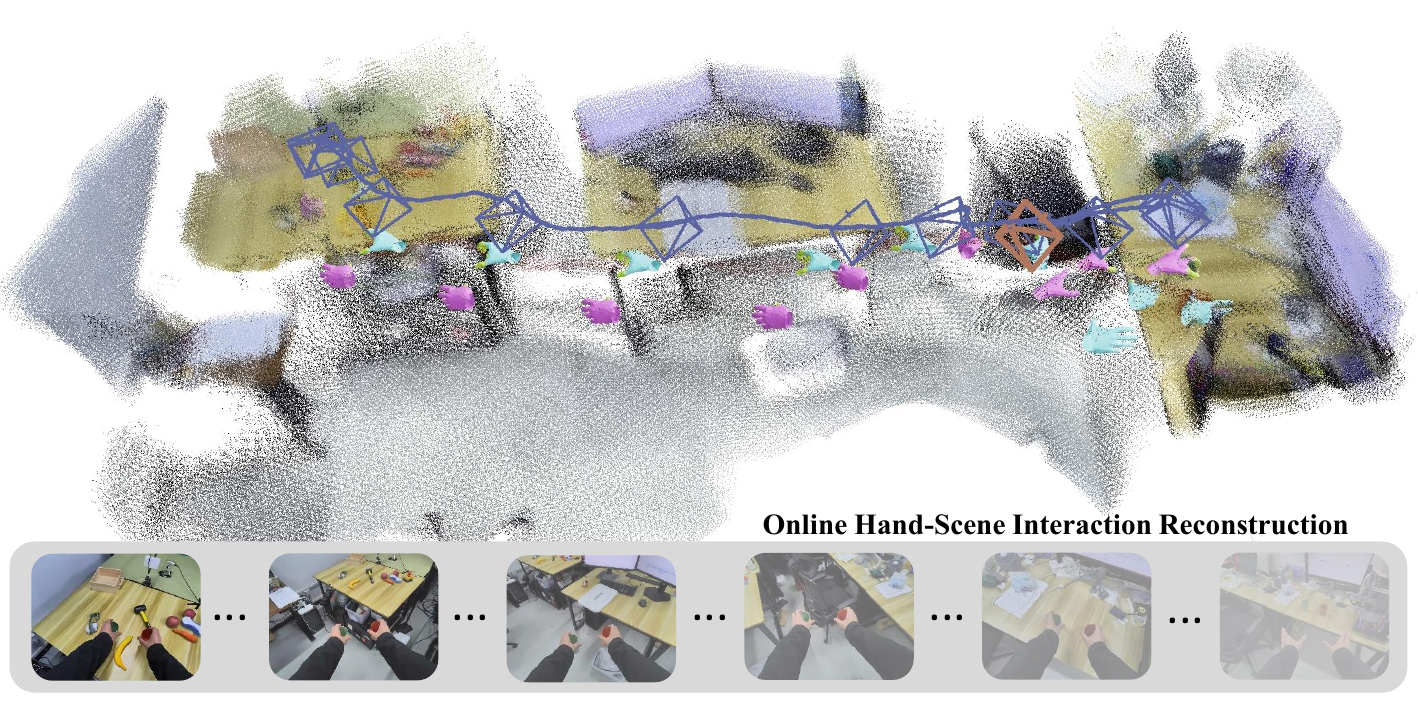}
    \caption{
    \methodname{} reconstructs camera motion, dense scene geometry, world-space bilateral MANO~\citep{mano} hands, and dense hand contact from egocentric video in a streaming manner. 
\textit{Top:} Reconstructed scene geometry accumulated from
multiple frames in a shared world coordinate system,
with the camera trajectory and sampled hand poses.
Left hands are shown in pink and right hands in cyan.
Predicted dense contacts are highlighted in green.
\textit{Bottom:} Sampled input egocentric frames.
    }
    \label{fig:teaser}

\end{figure*}

\begin{abstract}
\vspace{-5mm}
Egocentric videos offer an abundant source of hand-scene interactions for embodied learning, which requires recovering hands, camera, and scene consistently in a shared 3D world.
Existing methods focus on either camera-frame hands or world-space hand trajectories, without jointly recovering the scene pointmaps or interaction information.
Building a complete reconstruction from such specialized predictions therefore typically requires per-sequence optimization.
We propose \textbf{\methodname{}}, a unified feed-forward framework for online 4D hand-scene interaction reconstruction from egocentric monocular video.
To jointly recover these components, it maintains a recurrent scene representation that accumulates geometric evidence over time and provides a shared world space and scale.
We introduce hand-centric prompts that fuse pretrained hand priors and local scene context,
while MANO vertex queries attend to scene features to predict dense per-vertex contact. These designs enable a single network to jointly estimate camera motion, dense scene geometry, world-space bilateral MANO hands, and per-vertex hand contact.
Despite operating causally, \methodname{} achieves the lowest world-space hand joint error on ARCTIC among the evaluated methods, while offering faster inference than existing world-space hand reconstruction pipelines and competitive camera and scene reconstruction accuracy.
\end{abstract}

\section{Introduction}
\label{sec:intro}
\vspace{-5mm}
Egocentric videos provide abundant observations of hand--scene interactions for embodied learning. As human demonstrations become an increasingly important source of data for robot learning, extracting spatially grounded motion and interaction information from these videos is essential.
Recovering these interactions as structured 4D information requires jointly estimating camera motion, persistent dense scene geometry, world-space articulated hands, and dense hand-scene contact in a shared 3D world. Such a representation captures not only how the hands move, but also where and how they physically interact with the surrounding environment.

\begin{wrapfigure}{r}{0.6\textwidth}
    \vspace{-5mm}
    \centering
    \begin{minipage}[ht]{\linewidth}
        \centering
        \includegraphics[width=\linewidth]{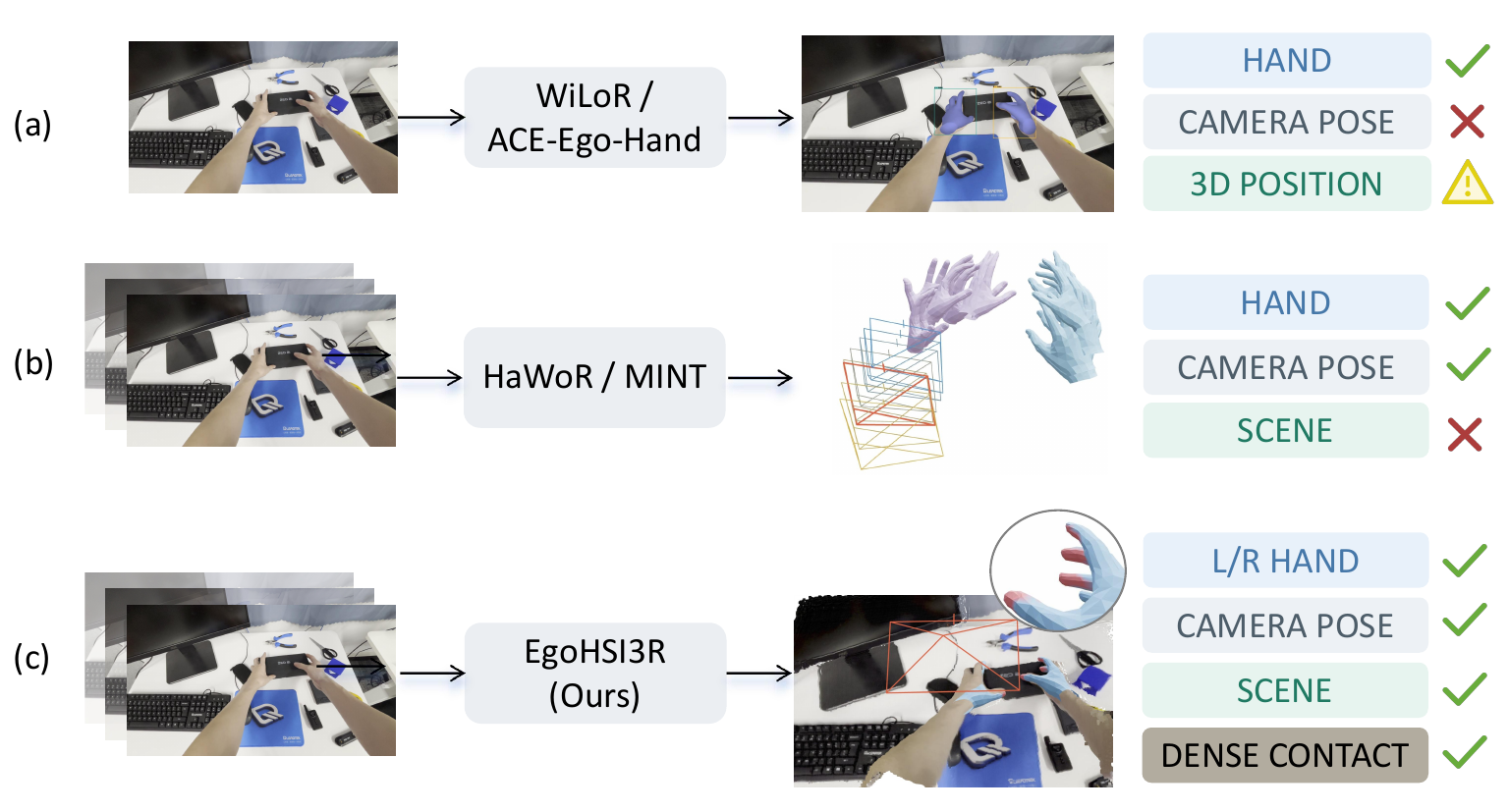}
        \vspace{-5mm}
        \caption{\textbf{Comparison of egocentric hand reconstruction paradigms.} (a) Camera-centric methods recover articulated hands in the camera frame, without camera motion and reliable 3D placement. (b) World-space methods additionally estimate camera motion to recover hand trajectories in world space, but do not reconstruct the scene. (c) \textbf{\methodname{}} jointly recovers bilateral hands, camera pose, scene geometry, and dense hand contact in a shared world frame within a single online, feed-forward model.}
        \label{fig:recent_method}
    \end{minipage}
    \vspace{-5mm}
\end{wrapfigure}

Recent progress in geometry foundation models has enabled recovering scene geometry from images or videos~\citep{dust3r,vggt,cut3r}. For hand reconstruction, Figure~\ref{fig:recent_method} summarizes two dominant paradigms in egocentric hand reconstruction. Camera-centric approaches recover detailed hand pose and shape from individual images or videos, but leave their predictions in the camera coordinate system without temporally consistent positions in the world~\citep{wilor,haptic,vidihand,aceegohand}. World-space approaches additionally estimate camera motion to recover hand trajectories in world space~\citep{hawor,dynhamr,stablehand,handflow, mint}. Both paradigms, however, provide only part of a complete hand-scene interaction reconstruction: they do not jointly recover the scene. 
Building a complete reconstruction from such specialized predictions therefore requires combining separate estimates and reconciling their coordinates, scales, and relative placement, typically through offline optimization. 
Such offline processing limits applications such as interactive robot teaching and human--robot collaboration, where scene and interaction estimates must be continuously updated as new observations arrive.

We therefore study online 4D hand-scene interaction reconstruction from egocentric video. Given only the frames up to the current time, our task is to incrementally recover \textbf{camera motion, persistent scene geometry, world-space trajectories of both MANO hands, and dense per-vertex contact in a shared world coordinate system.} 

This formulation introduces a challenge that does not arise in hand-only reconstruction: the hands and scene must be recovered in a consistent 3D configuration. Independently reconstructed hand trajectories and scene geometry may each appear plausible, yet disagree in their relative position or depth, resulting in floating hands or hand-scene interpenetration. The task therefore requires modeling the spatial relationship between the hands and scene as part of the prediction itself, rather than simply resolving this inconsistency after reconstruction by sequence-specific alignment.

To address this challenge, we present \textbf{\methodname{}}, an online feed-forward framework that jointly models the hands, scene, and their contact within a shared recurrent representation. The recurrent scene representation accumulates geometric evidence and camera motion over time, providing a persistent world space and scale shared by the hands and scene. We introduce hand-centric prompts that fuse pretrained hand priors and local scene context, and inject them into the recurrent representation to recover world-space trajectories of both MANO hands conditioned on the evolving scene. To explicitly model physical interaction, MANO vertex queries attend to local scene features to predict dense hand contact under contact-aware supervision. As illustrated in Figure~\ref{fig:teaser}, \methodname{} processes each incoming frame with one recurrent update and jointly reconstructs camera motion, persistent scene geometry, world-space trajectories of both MANO hands, and dense per-vertex hand--scene contact without test-time optimization.

We train \methodname{} on multiple egocentric datasets and evaluate it across scene reconstruction, world-space hand reconstruction, and contact prediction. \methodname{} achieves state-of-the-art world-space hand reconstruction on ARCTIC~\citep{arctic}, reducing W-MPJPE to 108.80\,mm, a 15.79\,mm improvement over the strongest baseline. 
Ablation studies further verify that both hand priors and interaction-aware contact feedback contribute to improved world-space hand estimation.

In summary, our main contributions are:
\begin{itemize}
\vspace{-2mm}
\item We propose \textbf{\methodname{}}, an online feed-forward framework for 4D hand-scene interaction reconstruction from egocentric monocular video, jointly recovering camera motion, persistent scene geometry, world-space bilateral MANO hands, and dense hand-scene contact.

\item We design a unified recurrent scene representation that enables joint hand-scene estimation over time, together with hand-centric prompts and vertex-level contact queries for scene-conditioned bilateral hand reconstruction and dense contact modeling.

\item \methodname{} achieves state-of-the-art performance in world-space hand reconstruction, while retaining competitive camera pose estimation and scene reconstruction accuracy in an online feed-forward setting.

\end{itemize}

\section{Related Work}
\label{sec:related_works}

\subsection{Monocular 3D and 4D Hand Reconstruction}
\vspace{-2mm}
Single-image methods recover detailed 3D hands from unconstrained RGB images using parametric hand models and strong learned priors~\citep{hamer,hamba,handos,wilor}. Egocentric approaches further address severe truncation, distortion, and self-occlusion~\citep{wildhand,egoforce}. However, these methods mainly operate in the camera frame and process frames independently.

Video-based methods improve temporal consistency through motion priors, tracking, or generative modeling~\citep{hmp,haptic,omnihands,vidihand,aceegohand}. Recent systems further reconstruct hands in world coordinates by combining hand observations with estimated camera motion~\citep{hawor,dynhamr,stablehand,handflow}. MINT~\citep{mint} jointly predicts camera and hand trajectories from egocentric video, but does not maintain dense scene geometry or explicitly model hand--scene contact. In contrast, our method jointly reconstructs world-space hands, camera motion, scene geometry, and contact within a causal framework.

\subsection{Feed-Forward 3D/4D Reconstruction}
\vspace{-2mm}
Feed-forward geometry models directly infer pointmaps and camera parameters without per-scene optimization. DUSt3R and VGGT establish strong multi-view reconstruction paradigms~\citep{dust3r,vggt}, while Spann3R, SLAM3R, and CUT3R introduce spatial memory or recurrent state for sequential reconstruction~\citep{spann3r,slam3r,cut3r}. Recent methods further improve long-sequence scalability through bounded memory or relative coordinate representations~\citep{infinitevggt,lingbotmap,r3}.

Dynamic extensions reconstruct both scene geometry and motion using time-dependent pointmaps, tracking, or conditional queries~\citep{monst3r,st4rtrack,d4rt,any4d,fourrc}. Nevertheless, these models represent humans and hands only as generic dynamic geometry and do not recover articulated hand parameters or interaction states. 

\subsection{Joint Human-Scene Reconstruction}
\vspace{-2mm}
Joint human--scene reconstruction has been explored through optimization-based systems such as JOSH and TRAM ~\citep{josh,tram}, as well as feed-forward models including UniSH and Human3R~\citep{unish,human3r}. These approaches primarily focus on offline, third-view full-body SMPL reconstruction, leaving detailed hand articulation and hand–scene interaction underexplored. Our work addresses this gap.

\subsection{Dense Hand Contact and Physical Interaction Modeling}
\vspace{-2mm}
ContactPose provides paired hand pose and surface-contact annotations~\citep{contactpose}, while CPF and ContactOpt use contact constraints to refine hand--object configurations~\citep{cpf,contactopt}. HACO predicts dense hand contact directly from visual observations~\citep{haco}, and ContactPrompt explores zero-shot contact estimation using multimodal models~\citep{contactprompt}. Recent methods jointly estimate hand--object pose~\citep{egograsp,hopformer} or extend contact prediction toward physical force estimation~\citep{egophi}. Most existing approaches estimate contact from individual frames or assume an object mesh is already available. In our formulation, we treat contact as the connection between hand and scene.

\begin{figure*}[t]
    \centering
    \includegraphics[width=\textwidth]{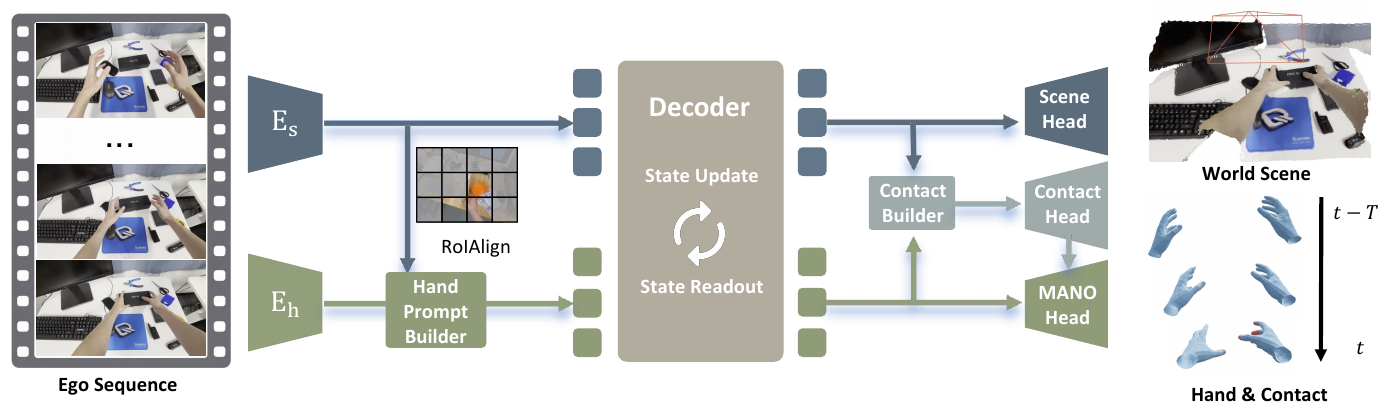}
    \vspace{-7mm}
    \caption{
    \textbf{Overview of \methodname{}.} Each incoming egocentric frame is encoded by the scene encoder $E_s$ into image tokens, and each detected hand is processed by the pretrained hand encoder $E_h$. The \textit{Hand Prompt Builder} fuses the hand prior with local scene context obtained via RoIAlign, producing scene-conditioned hand prompts. The decoder jointly processes scene tokens and hand prompts while reading from and updating a persistent scene state. The Scene Head predicts camera pose and world-space pointmaps, and the MANO Head reconstructs bilateral MANO parameters. The interaction-aware \textit{Contact Builder} combines decoded hand tokens with local scene features, which are then used by the Contact Head to predict per-vertex contact and refine hand reconstruction. All outputs are produced online in a shared world space without test-time optimization.
    }
    \vspace{-4mm}
    \label{fig:framework}
\end{figure*}

\section{Method}
\label{sec:method}

We address online 4D hand-scene interaction reconstruction by maintaining
a recurrent scene representation that is updated as each egocentric frame arrives.
The key mechanism is to integrate hand and dense contact estimation into recurrent scene reconstruction, allowing hands, scene geometry,
and their interaction to be recovered in a shared world space without
future observations or per-sequence optimization. 
We first formulate the streaming reconstruction task in Sec.~\ref{sec:problem} and summarize the
overall framework in Sec.~\ref{sec:framework}. We then describe the
\emph{Hand Prompt Builder} in Sec.~\ref{sec:prompt}, the
\emph{Interaction-Aware Contact Builder} in Sec.~\ref{sec:contact}, and the
training objectives in Sec.~\ref{sec:training}.

\subsection{Problem Formulation}
\label{sec:problem}
\vspace{-2mm}
Given a monocular egocentric RGB stream
$\{\mathbf{I}_t\}_{t=1}^{T}$, where
$\mathbf{I}_t\in\mathbb{R}^{H\times W\times3}$,
the problem is to use only the observed prefix $\mathbf{I}_{\leq t}$ at each
time $t$ to recover camera poses
$\mathbf{T}_t\in\mathrm{SE}(3)$, dense world-coordinate pointmaps
$\mathbf{X}_t\in\mathbb{R}^{H\times W\times3}$, and bilateral hand states
$(\boldsymbol{\Theta}_t^h,\boldsymbol{\beta}_t^h,
\boldsymbol{\tau}_t^h,\mathbf{c}_t^h)$ for
$h\in\{\mathrm{R},\mathrm{L}\}$.
For each hand,
$\boldsymbol{\Theta}_t^h\in\mathrm{SO}(3)^{16}$ contains the rotations of the
$16$ MANO joints, $\boldsymbol{\beta}_t^h\in\mathbb{R}^{10}$ represents hand
shape, $\boldsymbol{\tau}_t^h\in\mathbb{R}^{3}$ denotes camera-coordinate
translation, and $\mathbf{c}_t^h\in[0,1]^{778}$ contains the contact
probability of each MANO vertex.

\subsection{Framework Overview}
\label{sec:framework}
\vspace{-2mm}
To solve the streaming reconstruction task, we propose \methodname{}, an
online feed-forward framework that jointly reconstructs camera motion, scene
geometry, bilateral hands, and dense contact using a recurrent
representation, as illustrated in Fig.~\ref{fig:framework}. We instantiate this representation with
CUT3R~\citep{cut3r}, whose persistent state accumulates scene geometry and camera motion over time, and use WiLoR~\citep{wilor} to extract crop-based hand priors. Two components connect these representations. The \emph{Hand Prompt Builder} injects scene-conditioned hand priors into
the recurrent reconstruction process, while the
\emph{Interaction-Aware Contact Builder} associates decoded hand features
with local scene evidence to predict dense contact.

\subsection{Hand Prompt Builder}
\label{sec:prompt}
\vspace{-2mm}
Visual prompting provides an efficient way to integrate parametric human priors into recurrent scene reconstruction~\citep{human3r}. We adapt this principle to bilateral hands and condition each hand prompt on its local scene context.

For frame $\mathbf{I}_t$, the scene encoder produces patch tokens
$\mathbf{F}_t\in\mathbb{R}^{N\times d_s}$. An online hand detector localizes
each visible hand with a bounding box $\mathbf{b}_t^h$. We crop the detected
region and use the hand encoder to extract a structured prior
$\mathbf{A}_t^h\in\mathbb{R}^{M\times d_h}$, where $M=18$. Its first $16$
tokens encode articulated pose cues, while the remaining two summarize shape
and global translation.

A hand crop provides detailed articulation cues but removes much of the
surrounding geometry needed to place the hand in the scene. We therefore use
RoIAlign to extract local scene features
$\mathbf{R}_t^h=\operatorname{RoIAlign}(\mathbf{F}_t,\mathbf{b}_t^h)$ and
pool them into a context vector $\mathbf{s}_t^h$. The projected hand prior is
combined with a learned handedness embedding, and a feature-wise gate controls
the injection of local scene context:

\begin{equation}
\label{eq:hand_prompt}
\mathbf{P}_t^h
=
\phi_{\mathrm{mix}}
\left(
\mathbf{q}_t^h+\mathbf{g}_t^h\odot\mathbf{s}_t^h
\right),
\quad
\mathbf{q}_t^h=\phi_{\mathrm{h}}(\mathbf{A}_t^h)
+\mathbf{e}_{\mathrm{side}}^h,
\quad
\mathbf{g}_t^h=\sigma\!\left(
\phi_{\mathrm{g}}([\mathbf{q}_t^h;\mathbf{s}_t^h])
\right).
\end{equation}

This construction preserves the structured hand prior while adapting it to
the hand's local scene context. We assign the resulting prompt tokens the
spatial position of the bounding-box center. An absent hand is represented by
learned mask tokens, maintaining a fixed bilateral token layout.

We append the two hand prompts to CUT3R's camera and scene tokens, where self-attention and
bidirectional cross-attention enable the exchange of hand,
scene, and historical information. The decoder consequently produces
scene-aware hand tokens $\widetilde{\mathbf{P}}_t^h$ for contact and MANO
prediction, together with hand-conditioned scene tokens
$\widetilde{\mathbf{F}}_t$ for dense reconstruction.

\subsection{Interaction-Aware Contact Builder}
\label{sec:contact}
\vspace{-2mm}
The decoded hand tokens provide a compact joint-level representation, whereas
contact is defined densely over the MANO surface. To bridge this difference,
we first fuse $\widetilde{\mathbf{P}}_t^h$ with the original hand prior
$\mathbf{A}_t^h$. We then propagate the first $J=16$ pose-token features to
the $V=778$ MANO vertices using the fixed MANO skinning matrix
$\mathbf{W}\in\mathbb{R}^{V\times J}$. For vertex $i$, its query feature is

\begin{equation}
\label{eq:vertex_query}
\mathbf{v}_{t,i}^h
=
\operatorname{LN}\!\left(
\sum_{j=1}^{J}W_{ij}\mathbf{u}_{t,j}^h
+\overline{\mathbf{u}}_{t,\mathrm{g}}^h
+\mathbf{e}_{i}^{\mathrm{vert}}
+\mathbf{e}_{\mathrm{side}}^h
\right),
\end{equation}

where $\mathbf{u}_{t,j}^h$ is the fused feature of joint token $j$,
$\overline{\mathbf{u}}_{t,\mathrm{g}}^h$ summarizes the two non-pose tokens,
and $\mathbf{e}_{i}^{\mathrm{vert}}$ is a learned vertex embedding. The MANO
skinning weights preserve the correspondence between articulated joints and
surface vertices without introducing an additional learned topology mapping.

To obtain scene evidence, we apply RoIAlign to the decoded scene tokens using
the same hand box, yielding local context $\mathbf{C}_t^h$. Each vertex query
attends to this context through multi-head cross-attention, followed by a
feed-forward block and a binary classifier. The contact probability is
therefore computed as
$\widehat{c}_{t,i}^h=\sigma(\phi_{\mathrm{c}}(
\operatorname{FFN}(\mathbf{v}_{t,i}^h+
\operatorname{MHA}(\mathbf{v}_{t,i}^h,\mathbf{C}_t^h,\mathbf{C}_t^h))))$.
Restricting the context to hand neighborhood suppresses irrelevant
background features while retaining nearby object surfaces.

Beyond contact classification, the interaction-aware vertex features are
averaged and projected to a residual update $\Delta\mathbf{P}_t^h$ for the
decoded hand tokens. With a learnable gate $\alpha$, we update them as
$\widetilde{\mathbf{P}}_t^{h\prime}=\widetilde{\mathbf{P}}_t^h+
\sigma(\alpha)\Delta\mathbf{P}_t^h$. The final decoded hand tokens are
concatenated with $\mathbf{A}_t^h$ and mapped to $16$ MANO rotations,
$10$ shape coefficients, and a 3D translation. We initialize the feedback
projection to zero and its gate near zero so that optimization starts from the pretrained reconstruction solution and gradually incorporates interaction
cues. In this way, the Contact Builder not only predicts contact but also closes the hand--scene reasoning loop by refining the reconstructed hand.

\subsection{Training Objectives}
\label{sec:training}

The complete objective combines scene reconstruction, MANO reconstruction,
and contact estimation supervision:

\begin{equation}
\label{eq:total_objective}
\mathcal{L}
=
\lambda_{\mathrm{rec}}\mathcal{L}_{\mathrm{rec}}
+\lambda_{\mathrm{mano}}\mathcal{L}_{\mathrm{mano}}
+\lambda_{\mathrm{contact}}\mathcal{L}_{\mathrm{contact}}.
\end{equation}

We retain the original objectives of the scene reconstruction backbone in
$\mathcal{L}_{\mathrm{rec}}$, including dense geometry, camera pose,
confidence, and RGB supervision. The MANO objective supervises joint
rotations, shape, and camera-coordinate translation, together with 3D joints
and vertices in both root-relative and camera coordinate systems.

Because contact vertices are sparse, $\mathcal{L}_{\mathrm{contact}}$ uses a
focal loss over valid MANO vertices. We further partition the hand surface
into six anatomical regions according to the dominant MANO skinning joints.
A region is positive if any of its vertices is in contact, leading to
$\mathcal{L}_{\mathrm{contact}}=\mathcal{L}_{\mathrm{focal}}+
\lambda_{\mathrm{part}}\mathcal{L}_{\mathrm{part}}$.

\begin{table}[t]
\vspace{-1mm}
\caption{\textbf{Evaluation of Camera-space Hand Motion Estimation}. \emph{$p$-free} denotes penalty-free evaluation, where failed hand
detections are not penalized.  }
\centering
\footnotesize
\renewcommand{\arraystretch}{1.0}
\setlength{\tabcolsep}{2.5pt}
\resizebox{\linewidth}{!}{%
\begin{tabular}{@{}lcccccc@{}}
\toprule
& \multicolumn{2}{c}{\textbf{3D Pose}} 
& \multicolumn{3}{c}{\textbf{Orient. \& Position}} 
& \textbf{Temporal} \\
\textbf{Method}
& MPJPE-p$\downarrow$ 
& PA-p$\downarrow$ 
& EPE$_{\mathrm{2D\text{-}p}}\downarrow$ 
& GO-p$\downarrow$ 
& CT-p$\downarrow$ 
& Jitter$\downarrow$ \\
\midrule

\multicolumn{7}{@{}l}{\textbf{ARCTIC}} \\
HaMeR~\citep{hamer}                   & 29.197 & 14.596 & 65.289 & 24.907 & 0.095 & 18.279 \\
WildHands~\citep{wildhand}            & 25.704 & 13.941 & 50.517 & 22.320 & 0.058 & 12.972 \\
WiLoR~\citep{wilor}                   & 22.012 & 11.873 & 71.527 & 17.358 & 0.075 & 24.091 \\
EgoForce$^{\ddagger}$~\citep{egoforce}& 22.388 & 14.322 & 61.761 & 21.073 & 0.069 & 24.158 \\
OmniHands~\citep{omnihands}           & 29.674 & 14.203 & 51.505 & 24.580 & 0.087 & 45.312 \\
HaWoR~\citep{hawor}                   & 45.357 & 26.375 & 158.062 & 43.325 & 0.149 & 19.789 \\
ViDiHand~\citep{vidihand}             & 21.668 & 9.821  & {12.407} & 14.642 & 0.047 & \underline{3.183} \\
ACE-Ego-Hand~\citep{aceegohand} 

& \underline{16.615} & 8.756 & \underline{11.014} & \bf{12.208} & \bf{0.028} & \bf{2.770} \\

\textbf{Ours (Online)}                & 19.706 & \underline{7.289}  & 54.131 & 16.295 & 0.047 & 13.938 \\
\textbf{Ours (Online, p-free)}       & \bf{16.413} & \bf{7.025} & \bf{10.844} & \underline{12.621} & \underline{0.030} & 13.938 \\
\midrule

\multicolumn{7}{@{}l}{\textbf{HOT3D}} \\
HaMeR~\citep{hamer}                   & 68.314 & 21.455 & 59.077 & 49.636 & 0.102 & 23.632 \\
WildHands~\citep{wildhand}            & 52.791 & 28.946 & 111.438 & 53.933 & 0.157 & 22.885 \\
WiLoR~\citep{wilor}                   & 30.966 & 19.980 & 72.978 & 25.746 & 0.098 & 17.976 \\
EgoForce$^{\ddagger}$~\citep{egoforce}& 43.960 & 25.709 & 83.521 & 37.144 & 0.130 & 38.342 \\
OmniHands~\citep{omnihands}           & 63.281 & 22.682 & 68.437 & 49.120 & 0.133 & 69.510 \\
HaWoR~\citep{hawor}                   & 71.396 & 66.031 & 327.294 & 79.350 & 0.262 & 23.872 \\
ViDiHand~\citep{vidihand}             & 21.514 & 11.383 & \underline{14.953} & 15.829 & \underline{0.040} & \underline{3.741} \\
ACE-Ego-Hand~\citep{aceegohand} 
                                      & \bf{13.535} & \underline{6.768} & \bf{6.067} & \bf{8.362} & \bf{0.026} & \bf{3.426} \\

\textbf{Ours (Online)}                & 23.409 & 7.041  & 55.885 & 18.172 & 0.059 & 14.154 \\
\textbf{Ours (Online, p-free)}        & \underline{16.160} & \bf{6.312} & 15.470 & \underline{10.334} & 0.041 & 14.154 \\
\bottomrule
\end{tabular}%
}
\label{tab:hand_reconstruction}
\vspace{-2mm}
\end{table}

\begin{figure}[t]
    \centering
    \includegraphics[width=0.9\linewidth]{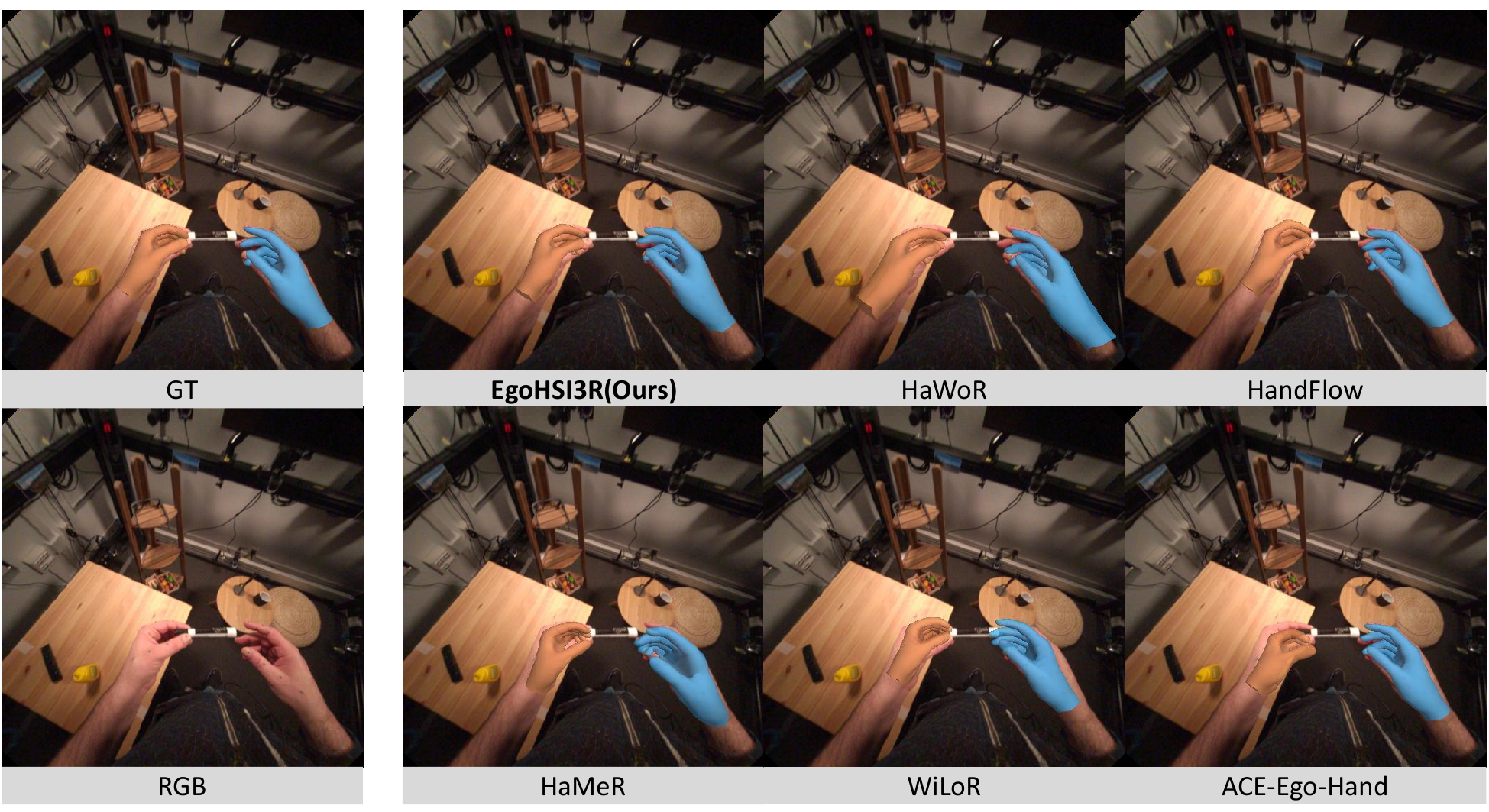}
    \caption{\textbf{Qualitative comparison of camera-space hand reconstruction.} The predicted hand meshes are overlaid on the input egocentric frames. Our method produces accurate articulated hand poses and consistent alignment with the observed images.}
    \label{fig:cam-space}
\end{figure}

\begin{table}[t]
\vspace{-5mm}
\caption{\textbf{Evaluation of World-space Hand Motion Estimation.}}
\centering
\footnotesize
\renewcommand{\arraystretch}{1.0}
\setlength{\tabcolsep}{2.5pt}
\resizebox{\linewidth}{!}{%
\begin{tabular}{@{}lcccc@{}}
\toprule
\textbf{Method} 
& \textbf{W-MPJPE $\downarrow$[mm]} 
& \textbf{WA-MPJPE $\downarrow$[mm]} 
& \textbf{AccEr $\downarrow$[m/s$^2$]} 
& \textbf{MRRPE $\downarrow$[mm]} \\
\midrule
HaMeR-SLAM~\citep{hamer}      & 232.48 & 59.35 & 10.82 & 163.46 \\
WiLoR-SLAM~\citep{wilor}      & 183.72 & 50.03 & 8.80  & 122.42 \\
HMP-SLAM~\citep{hmp}          & 166.21 & 41.62 & 7.13  & 131.33 \\
\midrule
Dyn-HaMR~\citep{dynhamr}      & 734.31 & 83.63 & \underline{6.05} & 189.09 \\
HaWoR~\citep{hawor}           & 171.61 & 51.98 & 11.35 & \underline{112.36} \\
StableHand~\citep{stablehand} & \underline{124.59} & \underline{32.66} & \bf{4.67} & \bf{93.71} \\
\textbf{Ours (Online)}        & \bf{108.80} & \bf{25.82} & 16.91 & 120.77 \\
\bottomrule
\end{tabular}%
}
\label{tab:world_hand}
\vspace{-2mm}
\end{table}

\section{Experiments}
\label{sec:exp}

We use six egocentric hand--object datasets with complementary supervision. The model is trained with the
available scene geometry, camera, MANO, and contact annotations from each
dataset. We evaluate camera-space hand reconstruction on ARCTIC~\citep{arctic} and HOT3D~\citep{hot3d},
world-space hand reconstruction on ARCTIC, contact estimation on ARCTIC and
DexYCB~\citep{dexycb}, and egocentric scene reconstruction on H2O~\citep{h2o}. The dataset composition and runtime comparison are provided in Tab.~\ref{tab:app_datasets} and Tab.~\ref{tab:app_efficiency} in the Appendix.

\vspace{-3mm}\subsection{Camera-Space Hand Reconstruction}
\vspace{-3mm}
\paragraph{Baselines.}
We compare with representative monocular and egocentric hand reconstruction
methods, including HaMeR~\citep{hamer}, WildHands~\citep{wildhand}, WiLoR~\citep{wilor}, EgoForce~\citep{egoforce}, OmniHands~\citep{omnihands}, HaWoR~\citep{hawor}, ViDiHand~\citep{vidihand}, and ACE-Ego-Hand~\citep{aceegohand}. 

\vspace{-4mm}
\paragraph{Evaluation Protocols.}
We evaluate camera-space hand reconstruction on ARCTIC~\citep{arctic} and HOT3D~\citep{hot3d} following the evaluation protocols used by the compared methods. We report MPJPE-p, PA-p, 2D projection error
EPE$_{\mathrm{2D\text{-}p}}$, global orientation error (GO-p), camera
translation error (CT-p), and temporal jitter. Lower values are better for all
metrics. A GT hand is evaluated if at least one projected joint lies inside
the image; missing or 2D-unmatched predictions are penalized even with GT boxes.
Since causal online methods cannot use future observations to complete missing
motion, we also report \emph{$p$-free} results on successfully matched predictions
to measure conditional reconstruction accuracy. Note that this is a matched-only
diagnostic, not an evaluation over GT-visible hands.

\vspace{-4mm}
\paragraph{Results.}
As shown in Table~\ref{tab:hand_reconstruction}, our $p$-free setting achieves the best PA-p on both datasets, with $7.025$ and $6.312$ mm, respectively. It also achieves the best MPJPE-p on ARCTIC, the second-best MPJPE-p on HOT3D, and the second-best GO-p on both datasets. This demonstrates high conditional reconstruction accuracy, while the standard results additionally reflect missed or unmatched predictions under partial visibility.
Figure~\ref{fig:cam-space} provides qualitative comparisons on HOT3D, where our predictions show accurate alignment with the observed hand configurations. Our temporal jitter remains higher than methods designed specifically for motion smoothing, indicating room for improving temporal stability without sacrificing online operation.

\subsection{World-Space Hand Reconstruction}

\vspace{-2mm}
\paragraph{Baselines.}
We compare against single-frame hand estimators coupled with SLAM~\citep{droid}, including
HaMeR-SLAM, WiLoR-SLAM, and HMP-SLAM, as well as world-space hand motion
methods Dyn-HaMR~\citep{dynhamr}, HaWoR~\citep{hawor}, and StableHand~\citep{stablehand}.

\vspace{-3mm}
\paragraph{Evaluation Protocols.}
We evaluate world-space hand motion on ARCTIC~\citep{arctic}. W-MPJPE measures absolute
world-coordinate joint error, WA-MPJPE evaluates the trajectory after global
alignment, AccEr measures temporal acceleration error, and MRRPE evaluates the
relative placement of the two hands. Lower values indicate better performance.

\begin{figure}[t]
    \centering
    \includegraphics[width=\linewidth]{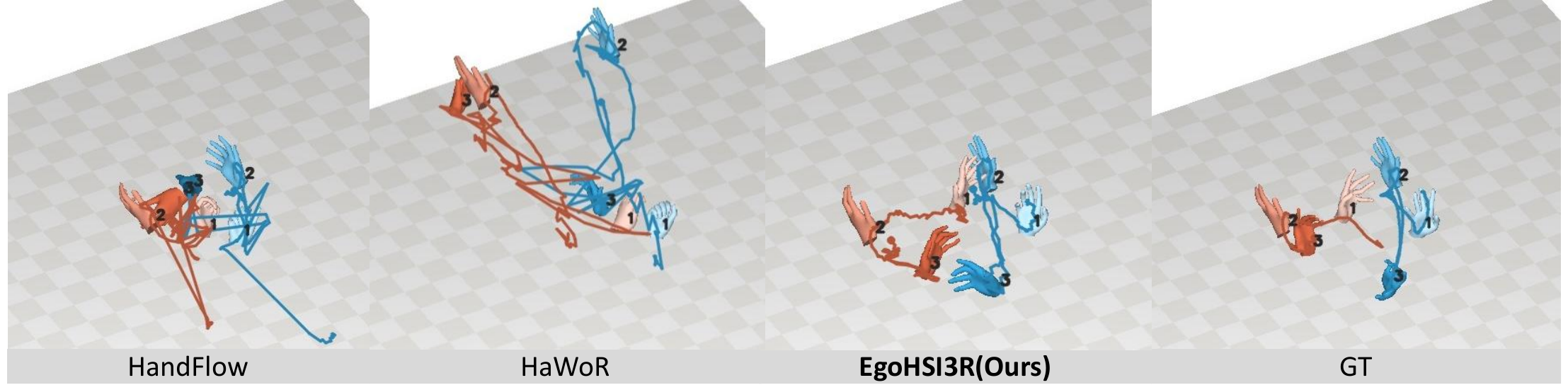}
    \caption{\textbf{Qualitative comparison of world-space hand reconstruction.} Existing methods~\citep{handflow, hawor} exhibit inaccurate hand trajectories and spatial inconsistencies, while \methodname{} produces world-space hand motions closer to the ground truth. The numbers indicate timestamps at 0.83s, 2.50s, and 4.17s, respectively.}
    \label{fig:world-space}
\end{figure}

\vspace{-3mm}
\paragraph{Results.}
Table~\ref{tab:world_hand} shows that our online model achieves the best
W-MPJPE and WA-MPJPE, reducing them to $108.80$ mm and $25.82$ mm. Compared
with the strongest baseline, StableHand, this corresponds to improvements of
$15.79$ mm and $6.84$ mm, respectively. Figure~\ref{fig:world-space} provides
qualitative comparisons on ARCTIC, where \methodname{} maintains temporally
consistent world-space hand poses and recovers more accurate world-space hand
motions.  The higher AccEr and MRRPE are partly attributable to causal inference,
which precludes future-frame smoothing and sequence-level correction of bilateral trajectories under intermittent visibility.

\begin{table}[t]
\centering
\caption{
\textbf{Contact estimation} on ARCTIC~\citep{arctic} and DexYCB~\citep{dexycb}.
We report per-vertex contact precision, recall, and F1-score.
}
\small
\setlength{\tabcolsep}{6pt}
\renewcommand{\arraystretch}{1.05}

\begin{tabular}{l|ccc|ccc}
\toprule
& \multicolumn{3}{c|}{\textbf{ARCTIC}}
& \multicolumn{3}{c}{\textbf{DexYCB}} \\
\textbf{Method}
& Prec.$\uparrow$ & Rec.$\uparrow$ & F1$\uparrow$
& Prec.$\uparrow$ & Rec.$\uparrow$ & F1$\uparrow$ \\
\midrule

HACO~\citep{haco}
& 43.609 & \bf{ 65.181}&\bf{  48.917}
& 43.993 & \bf{61.304} & \bf{49.115} \\

\textbf{Ours}
& \bf{52.152} & 45.015 & 44.298
& \bf{51.619} & 50.376 & 48.197 \\

\bottomrule
\end{tabular}
\label{tab:contact}
\end{table}

\subsection{Contact Evaluation}
\paragraph{Baselines.}
We compare with HACO~\citep{haco}, a representative method for dense hand--object contact
estimation on the MANO surface.

\vspace{-4mm}
\paragraph{Evaluation Protocols.}
We evaluate binary contact prediction on ARCTIC~\citep{arctic} and DexYCB~\citep{dexycb} using the standard 778-vertex MANO topology. Precision, recall, and F1-score are computed over valid annotated hand vertices with a threshold of 0.5; higher values indicate better performance.

\vspace{-4mm}
\paragraph{Results.}
As reported in Table~\ref{tab:contact}, our method improves contact precision
over HACO on both ARCTIC and DexYCB, reaching $52.152$ and $51.619$,
respectively. Its lower recall leads to lower F1-scores, suggesting that the
current model produces more conservative contact predictions. These results
show improved reliability for predicted contact vertices while leaving recall
as the main direction for further improvement.

\subsection{Egocentric 3D Reconstruction}

\begin{table}[t]
\caption{\textbf{Evaluation of Egocentric 3D Reconstruction} on the H2O~\citep{h2o} dataset.}
\centering
\small
\renewcommand{\arraystretch}{1.}
\renewcommand{\tabcolsep}{2.5pt}
\resizebox{\linewidth}{!}{
\begin{tabular}{@{}clccccccc@{}}
\toprule
 & \textbf{Method} & \textbf{Optim.} & \textbf{Onl.} & {Abs Rel $\downarrow$} & {$\delta<1.25$ $\uparrow$} & {ATE $\downarrow$[mm]} & {RPEt $\downarrow$[mm]} & {RPEr $\downarrow$[°]} \\
\midrule
 & EgoMono4D~\citep{egomono4d} & \checkmark & $\times$ & 0.049 & \underline{0.976} & 7.149 & 2.606 & 0.781 \\
 & ReViV~\citep{reviv} & $\times$  & $\times$& 0.260 & 0.608 & \bf{2.383} & \bf{0.865} & \underline{0.075} \\
 \midrule
 & TTT3R~\citep{ttt3r} & $\times$ & \checkmark & \underline{0.045} & 0.965 & 3.578 & 1.271 & \bf{0.074} \\
 & \bf{Ours} & $\times$ & \checkmark & \bf{0.027} & \bf{0.993} & \underline{2.573} & \underline{0.923} & 0.079 \\
\bottomrule
\end{tabular}
}
\label{tab:video_pose}
\end{table}

\paragraph{Baselines.}
We compare with EgoMono4D~\citep{egomono4d} and ReViV~\citep{reviv} as offline reconstruction baselines, and
with TTT3R~\citep{ttt3r} as an online recurrent reconstruction method. 

\vspace{-4mm}
\paragraph{Evaluation Protocols.}
We evaluate scene depth and camera tracking on H2O~\citep{h2o}. Depth quality is measured
by absolute relative error (Abs Rel) and threshold accuracy
$\delta<1.25$. Camera trajectories are evaluated using absolute trajectory
error (ATE) and relative translation and rotation errors (RPEt and RPEr).
Abs Rel, ATE, RPEt, and RPEr are minimized, whereas $\delta<1.25$ is
maximized.

\vspace{-4mm}
\paragraph{Results.}
As shown in Table~\ref{tab:video_pose}, our method achieves the best depth
accuracy, with an Abs Rel of $0.027$ and $\delta<1.25$ of $0.993$. It also
ranks second in ATE and RPEt, closely approaching the offline ReViV method,
while maintaining causal online inference without test-time optimization.
Its RPEr of $0.079^{\circ}$ is comparable to the best online result.

\subsection{Ablation Study}
Table~\ref{tab:model_design_ablation} examines the contributions of hand fine-tuning, contact prediction, and contact feedback. 

Fine-tuning WiLoR substantially improves hand reconstruction over the base model, reducing ARCTIC MPJPE from 22.916\,mm to 19.952\,mm and W-MPJPE from 127.855\,mm to 114.067\,mm. Because the WiLoR prior was originally trained on right hands, fine-tuning substantially improves bilateral hand reconstruction.

Without contact feedback, the jointly trained contact head achieves F1 scores of 43.349 on ARCTIC and 47.657 on DexYCB. Its ARCTIC and HOT3D MPJPE are slightly higher than those of the hand-only baseline, whereas ARCTIC W-MPJPE improves modestly from 114.067 to 113.296 mm. This mixed outcome may reflect optimization instability introduced by jointly training the hand reconstruction and contact prediction objectives.

With contact feedback enabled, ARCTIC and HOT3D MPJPE decrease by 0.570 and 0.705 relative to the no-feedback variant, while W-MPJPE decreases by 4.495\,mm. Contact F1 also improves on both datasets. Since the final two variants share the same warm-up and joint training procedure, these gains support the value of incorporating contact cues into the reconstruction process.

\begin{table}[t]
\vspace{-5mm}
\centering
\caption{\textbf{Ablation study.} The two variants with a contact head use the same contact-head warm-up and joint training procedure.}
\label{tab:model_design_ablation}
\resizebox{\linewidth}{!}{%
\begin{tabular}{lccc|ccccc}
\toprule
Variant & WiLoR FT & Contact head & Contact feedback
& \shortstack{ARCTIC\\MPJPE [mm]$\downarrow$}
& \shortstack{HOT3D\\MPJPE [mm] $\downarrow$}
& \shortstack{ARCTIC\\W-MPJPE [mm] $\downarrow$}
& \shortstack{ARCTIC\\F1 $\uparrow$}
& \shortstack{DexYCB\\F1 $\uparrow$} \\
\midrule
Base model
& -- & -- & -- & 22.916 & 28.974 & 127.855 & -- & -- \\
Hand-only baseline
& $\checkmark$ & -- & -- & 19.952 & 23.653 & 114.067 & -- & -- \\
Without contact feedback
& $\checkmark$ & $\checkmark$ & -- & 20.276 & 24.114 & 113.296 & 43.349 & 47.657 \\
\textbf{\methodname{} (full)}
& $\checkmark$ & $\checkmark$ & $\checkmark$
& \textbf{19.706} & \textbf{23.409} & \textbf{108.801}
& \textbf{44.298} & \textbf{48.197} \\
\bottomrule
\end{tabular}%
}
\vspace{-5mm}
\end{table}

\section{Conclusion}
\label{sec:Conclusion}

We presented \textbf{EgoHSI3R}, an online feed-forward framework for unified 4D hand–scene interaction reconstruction from egocentric video. 
It maintains a recurrent scene representation that provides a shared world space for hand and scene reconstruction, while scene-conditioned hand prompts and interaction-aware contact modeling couple hand reconstruction with the evolving scene.
Together, these designs enable joint reconstruction of camera motion, persistent scene geometry, world-space bilateral MANO hands, and dense hand-scene contact without future observations or test-time optimization.
Experiments across multiple egocentric benchmarks demonstrate state-of-the-art world-space hand reconstruction while retaining competitive camera and scene reconstruction accuracy. These results highlight the benefit of jointly modeling hands, scene geometry, and physical interaction within a unified recurrent representation for egocentric interaction reconstruction.

\newpage

\subsection*{AI use statement}
We used AI tools only to improve the readability of the writing and to assist with identifying potential bugs in the code. All research ideas, methods, experimental results, and conclusions were developed and verified by the authors. The authors reviewed all AI-assisted content and take full responsibility for the final manuscript and code.

\subsection*{Reproducibility statement}
We provide the experimental settings and implementation details to facilitate reproducibility in Sec.~\ref{sec:method} and Sec.~\ref{sec:exp}, including the model architecture, training objectives, datasets, and evaluation protocols. We are also preparing the code for public release.

\bibliography{iclr2027_conference}
\bibliographystyle{iclr2027_conference}

\newpage

\appendix

\providecommand{\methodname}{Hand3R}
\providecommand{\appendixfigdir}{figures}

\section{Appendix}
\label{sec:appendix}

\subsection{Comparison with prior methods.}

Table~\ref{tab:method_compare} summarizes the capabilities of existing methods. Camera-space hand methods focus on articulated hand reconstruction, while world-space approaches additionally recover camera motion, often relying on external SLAM systems. Recent unified methods extend reconstruction to multiple components but do not explicitly model dense hand contact. In contrast, EgoHSI3R jointly reconstructs world-space hands, camera motion, scene geometry, and dense contact in a single causal, feed-forward framework.

\begin{table}[htbp]
\centering
\caption{Comparison with related methods. Online denotes causal inference, and FF denotes feed-forward inference. $^{*}$ indicates camera poses obtained from external estimation systems.}
\resizebox{\linewidth}{!}{%
\begin{tabular}{l|cc|cccc}
\toprule
\textbf{Method} &
\textbf{Online} & \textbf{FF} &
\textbf{World Hand} & \textbf{Camera} &
\textbf{Scene} & \textbf{Contact} \\
\midrule
WiLoR~\citep{wilor}
& -- & \checkmark
& $\times$ & $\times$ & $\times$ & $\times$ \\
HaPTIC~\citep{haptic}
& $\times$ & \checkmark
& $\times$ & $\times$ & $\times$ & $\times$ \\
HaWoR~\citep{hawor}
& $\times$ & $\times$
& \checkmark & $\checkmark^{*}$ & $\times$ & $\times$ \\
Dyn-HaMR~\citep{dynhamr}
& $\times$ & $\times$
& \checkmark & $\checkmark^{*}$ & $\times$ & $\times$ \\
StableHand~\citep{stablehand}
& $\times$ & $\times$
& \checkmark & $\checkmark^{*}$ & $\times$ & $\times$ \\
HandFlow~\citep{handflow}
& $\times$ & $\times$
& \checkmark & $\checkmark^{*}$ & $\times$ & $\times$ \\
ViDiHand~\citep{vidihand}
& $\times$ & \checkmark
& $\times$ & $\times$ & $\times$ & $\times$ \\
ACE-Ego-Hand~\citep{aceegohand}
& $\times$ & \checkmark
& $\times$ & $\times$ & $\times$ & $\times$ \\
MINT~\citep{mint}
& $\times$ & \checkmark
& \checkmark & \checkmark & $\times$ & $\times$ \\
ReViV~\citep{reviv}
& $\times$ & \checkmark
& \checkmark & \checkmark & \checkmark & $\times$ \\
HACO~\citep{haco}
& -- & \checkmark
& $\times$ & $\times$ & $\times$ & \checkmark \\
\midrule
\textbf{Ours}
& \checkmark & \checkmark
& \checkmark & \checkmark & \checkmark & \checkmark \\
\bottomrule
\end{tabular}
}
\label{tab:method_compare}
\end{table}

\subsection{Datasets and Data Labeling}
\label{sec:app_datasets}
As shown in Table~\ref{tab:app_datasets}, we train on six sources: HOT3D~\citep{hot3d}, ARCTIC~\citep{arctic},
TACO~\citep{taco}, H2O~\citep{h2o}, HOI4D~\citep{hoi4d}, and
DexYCB~\citep{dexycb}. Each processed frame provides RGB input and available
scene, camera, hand, and contact supervision, with explicit validity masks.
Splits are defined at the recording level: windows, frames, and camera streams
from the same recording remain in the same split.

\paragraph{Coordinate and validity conventions.}
The processed representation uses meters and OpenCV camera axes
($+x$ right, $+y$ down, $+z$ forward), with separate right/left MANO slots.
Intrinsics and bounding boxes are transformed consistently with image
rectification, resizing, cropping, and padding. Native side-specific MANO
layers preserve left/right geometry. Parameter validity, visible-hand
validity, and contact validity are distinguished, so unavailable supervision
is not silently converted into a negative label. Camera poses are exported
as camera-to-world transforms in the dataset's reference frame; the temporal
pose is fixed within each DexYCB camera stream.

\begin{table}[htbp]
\centering
\caption{Supervision available in the processed training sources. A star
marks generated supervision: rendered/completed depth or geometry-derived
contact. Availability does not imply that every frame or hand is valid.}
\label{tab:app_datasets}
\small
\setlength{\tabcolsep}{5pt}
\begin{tabular}{@{}lccccc@{}}
\toprule
Dataset & RGB & Depth & Pose & Hand & Contact \\
\midrule
HOT3D (Aria) & $\checkmark$ & $\checkmark^{*}$ & $\checkmark$ & $\checkmark$ & $\checkmark^{*}$ \\
ARCTIC & $\checkmark$ & $\checkmark^{*}$ & $\checkmark$ & $\checkmark$ & $\checkmark^{*}$ \\
TACO & $\checkmark$ & $\checkmark$ & $\checkmark$ & $\checkmark$ & $\checkmark^{*}$ \\
H2O & $\checkmark$ & $\checkmark$ & $\checkmark$ & $\checkmark$ & $\checkmark^{*}$ \\
HOI4D & $\checkmark$ & $\checkmark$ & $\checkmark$ & $\checkmark$ & $\checkmark^{*}$ \\
DexYCB & $\checkmark$ & $\checkmark$ & $\checkmark$ & $\checkmark$ & $\checkmark^{*}$ \\
\bottomrule
\end{tabular}
\end{table}

\paragraph{Depth labeling.}
ARCTIC depth is camera-$z$ depth rendered from the annotated hand and
articulated-object meshes; background without mesh support is invalid.
HOT3D uses hybrid LingBot-Depth completion: the foreground branch receives
native mesh depth together with sparse SLAM prompts, while the static
background branch receives SLAM-only prompts. Foreground/background regions
are selected using the exported segmentation and combined into the final
depth map. The stored source-dependent confidence map is not used to
reweight the final HOT3D depth supervision in these runs. TACO, H2O, HOI4D,
and DexYCB use their processed camera-aligned sensor depth, with dataset
specific scale conversion and invalid-depth masks.

\paragraph{Contact labeling.}
For MANO vertex $\mathbf v$ and the relevant posed object surface
$\mathcal S$, the generated binary label is
\begin{equation}
y(\mathbf v)=\mathbb{1}\!\left[d(\mathbf v,\mathcal S)\leq 0.01\;\mathrm{m}\right].
\label{eq:app_contact_label}
\end{equation}
This 1 cm geometric labeling threshold is distinct from the 0.5 probability
threshold used to classify predicted contacts. ARCTIC uses the articulated
object; TACO uses the tool and target meshes; H2O uses the annotated object;
DexYCB uses the grasped object. HOI4D uses unsigned point-to-triangle distance
to valid posed object parts. HOT3D approximates the surface with deterministic
object-vertex and triangle-centroid samples, capped at 30,000 samples per
object. Contact is supervised only when the required hand and object
annotations are valid. These labels describe geometric proximity, rather
than force measurements or independently annotated physical contact.

\subsection{Data Splits and Evaluation Protocol}
\label{sec:app_splits}
We use a frozen six-source split defined over complete recordings, rather
than splitting frames or temporal windows at random. All windows of a
recording and all its camera streams remain in the same partition.
Table~\ref{tab:app_split_counts} gives the available processed sequences;
these counts are distinct from the evaluation-clip counts below.
Validation is used for checkpoint and inference-setting selection; test
data are reserved for final scoring. All evaluations follow our data split protocol. 

For camera-space evaluation, we adopt the metric definitions of ACE-Ego-Hand~\citep{aceegohand} on ARCTIC and HOT3D, using non-overlapping 81-frame clips at 30 FPS. We report MPJPE, PA-MPJPE, EPE2D, global orientation error, camera-space translation error, and jitter. 

For world-space evaluation, we adopt the released metric implementation of StableHand~\citep{stablehand} on 150-frame ARCTIC clips, reporting W-MPJPE, wrist-aligned WA-MPJPE, acceleration error, and mean relative root position error (MRRPE). Predictions and ground truth are independently expressed relative to their respective first-frame cameras, with no scale or hand-root alignment for W-MPJPE. 

For contact estimation, we follow the per-hand-frame macro-averaging convention of HACO~\citep{haco} on complete ARCTIC and DexYCB recordings. Precision, recall, and F1 are computed over the 778 MANO vertices per hand using a prediction threshold of 0.5, with undefined per-hand-frame scores set to zero.

\begin{table}
\centering
\caption{Frozen processed-data split sizes. DexYCB counts camera streams; all other rows count processed sequences. }
\label{tab:app_split_counts}
\small
\setlength{\tabcolsep}{6pt}
\begin{tabular}{@{}lrrr@{}}
\toprule
Dataset & Train & Val & Test \\
\midrule
HOT3D & 103 & 11 & 22 \\
ARCTIC & 235 & 26 & 33 \\
TACO & 806 & 90 & 1,313 \\
H2O & 124 & 14 & 46 \\
HOI4D & 1,683 & -- & -- \\
DexYCB & 6,400 & 320 & 1,280 \\
\bottomrule
\end{tabular}
\end{table}

\subsection{Efficiency Analysis}
\label{sec:app_efficiency}
\paragraph{Baselines and measurement scope.}
We compare the final feedback model against a local WiLoR~\citep{wilor}+ViPE
pipeline, HaWoR~\citep{hawor}, and HandFlow~\citep{handflow}. WiLoR+ViPE is a local engineering composition, rather
than a reproduction of an unspecified SLAM backend. Measurements use one
NVIDIA H200, the same 150-frame ARCTIC input clips, and a 512-pixel input long
edge. Method-specific internal resizing is retained. One clip is used for
warmup and ten clips for measurement. Model loading and final visualization
are excluded. Necessary intermediate file I/O in HaWoR is included.

\begin{table}
\centering
\caption{Efficiency of the measured RGB-to-world pipelines. FPS is median
end-to-end throughput; memory is the maximum allocated CUDA memory across
measured clips in GiB. }
\label{tab:app_efficiency}
\small
\setlength{\tabcolsep}{4pt}
\begin{tabular}{@{}lrrr@{}}
\toprule
Method & Params (M) & FPS & Memory (GiB) \\
\midrule
WiLoR+ViPE & 708.954 & 7.545 & 6.142 \\
HaWoR & 1170.122 & 7.756 & 8.099 \\
HandFlow & 907.017 & 7.645 & 10.063 \\
\textbf{\methodname{}} & 1470.250 & \textbf{10.070} & 6.650 \\
\bottomrule
\end{tabular}

\end{table}

\paragraph{Results.}
As shown in Table~\ref{tab:app_efficiency}, the final model achieves 10.070 FPS end to end and 15.285 FPS for model
inference alone. These are different timing scopes: the former includes
image loading, preprocessing, detection, crop preparation, inference, and
output decoding. Model-only throughput is therefore not substituted into
the end-to-end comparison. Peak allocated memory is 6.650 GiB, below HaWoR
and HandFlow and above the local WiLoR+ViPE composition. The parameter count
includes the detector. For serial multi-process baselines, memory is the
maximum phase peak, rather than the sum of phase peaks.

\subsection{Training Loss Details}
\label{sec:app_losses}
We distinguish scene reconstruction, articulated hand reconstruction, and
contact prediction. The final training objective is
\begin{equation}
\begin{split}
\mathcal L={}&\mathcal L_{\mathrm{rec}}+\mathcal L_{\mathrm{MANO}}\\
 &+0.3\bigl(\mathcal L_{\mathrm{focal}}+0.1\mathcal L_{\mathrm{part}}\bigr),
\end{split}
\label{eq:app_total_loss}
\end{equation}
where
$\mathcal L_{\mathrm{rec}}=\mathcal L_{\mathrm{points}}+
\mathcal L_{\mathrm{pose}}+\mathcal L_{\mathrm{RGB}}$ comes from CUT3R.

\paragraph{MANO parameter supervision.}
Each valid hand has 16 joint rotation matrices $R$, shape coefficients
$\boldsymbol\beta$, and native MANO translation $\mathbf t$.
The side-specific differentiable MANO layer maps these parameters to camera
space joints $J$ and 778 vertices $V$. Define
$\ell(A,B)=|A|^{-1}\sum_k|A_k-B_k|$, a mean elementwise $L_1$ error.
The parameter loss is
\begin{equation}
\mathcal L_{\mathrm{param}}=
\ell(\hat R,R)+0.1\ell(\hat{\boldsymbol\beta},\boldsymbol\beta)
+\ell(\hat{\mathbf t},\mathbf t).
\label{eq:app_mano_param}
\end{equation}
Rotation supervision compares matrix entries, rather than axis-angle
vectors or geodesic distances. Native left and right MANO layers are used
without replacing one side with the other's geometry.

\paragraph{MANO joint and surface supervision.}
Let $\mathbf r=J_0$ and $\hat{\mathbf r}=\hat J_0$ denote the reference
and predicted wrists. Camera-space terms supervise absolute hand placement:
\begin{equation}
\mathcal L_{\mathrm{cam}}=\ell(\hat J,J)+\ell(\hat V,V).
\end{equation}
Wrist-relative terms separately constrain articulation and shape:
\begin{equation}
\begin{split}
\mathcal L_{\mathrm{rel}}={}&
\ell(\hat J-\hat{\mathbf r},J-\mathbf r)\\
&+\ell(\hat V-\hat{\mathbf r},V-\mathbf r).
\end{split}
\end{equation}
The same wrist is subtracted from every vertex or joint of a hand.
All four 3D terms have unit weight and use meter-valued coordinates;
they are not rescaled by the scene normalization factor $s$.
Parameter, joint, and vertex losses require image validity, a visible-hand
mask, and valid MANO parameters. Each term is averaged over its selected
entries and then over views. Batches with no valid hands contribute a
zero MANO loss.

\paragraph{MANO image-plane supervision.}
With the reference intrinsics, perspective projection is
\begin{equation}
\pi_K([x,y,z]^\top)=
[f_xx/z+c_x,\;f_yy/z+c_y]^\top.
\end{equation}
The joint and vertex reprojection terms are
\begin{equation}
\begin{aligned}
\mathcal L_{J2D}&=\ell\bigl(\pi_K(\hat J),\pi_K(J)\bigr),\\
\mathcal L_{V2D}&=\ell\bigl(\pi_K(\hat V),\pi_K(V)\bigr).
\end{aligned}
\end{equation}
These errors are measured in pixels. Only finite prediction/reference
pairs with both depths above 0.01 m are used, and the projected reference
point must lie strictly inside the image. Safe clamping before division
prevents invalid gradients near the camera plane. To avoid unstable
projection supervision before coarse geometry is learned, both terms
are gated by $g=\mathbb{1}[\operatorname{stopgrad}
(\ell(\hat V,V))<1.0]$, using the camera-space vertex loss.
The full final-stage hand objective is
\begin{equation}
\begin{split}
\mathcal L_{\mathrm{MANO}}={}&\mathcal L_{\mathrm{param}}
+\mathcal L_{\mathrm{cam}}+\mathcal L_{\mathrm{rel}}\\
&+0.005g\bigl(\mathcal L_{J2D}+\mathcal L_{V2D}\bigr).
\end{split}
\label{eq:app_mano_total}
\end{equation}

\paragraph{Dense contact and hand-part supervision.}
The contact branch predicts one logit for each of the 778 MANO vertices.
With binary target $y_v$ and probability $p_v=\sigma(z_v)$, define
$p_{t,v}=y_vp_v+(1-y_v)(1-p_v)$. The vertex loss is
\begin{equation}
\mathcal{L}_{\mathrm{focal}}=
-\frac{1}{778}\sum_{v=1}^{778}0.5(1-p_{t,v})^2\log p_{t,v}.
\end{equation}
The implementation evaluates this expression through numerically stable
binary cross-entropy with logits. Vertices are assigned to six coarse
regions (palm and five fingers) using dominant MANO skinning weights.
For region $\mathcal{V}_k$, the pooled logit and target are
\begin{equation}
\bar z_k=\log\sum_{v\in\mathcal{V}_k}\exp z_v-\log|\mathcal{V}_k|,
\qquad \bar y_k=\max_{v\in\mathcal{V}_k}y_v.
\end{equation}
The part loss averages binary cross-entropy over the six regions. Contact
loss is averaged over valid visible hand-frames, using contact validity,
image validity, and the hand mask. 

\subsection{Additional Qualitative Examples}
\label{sec:app_qualitative}
\begin{figure*}[h]
\centering
\includegraphics[width=\textwidth]{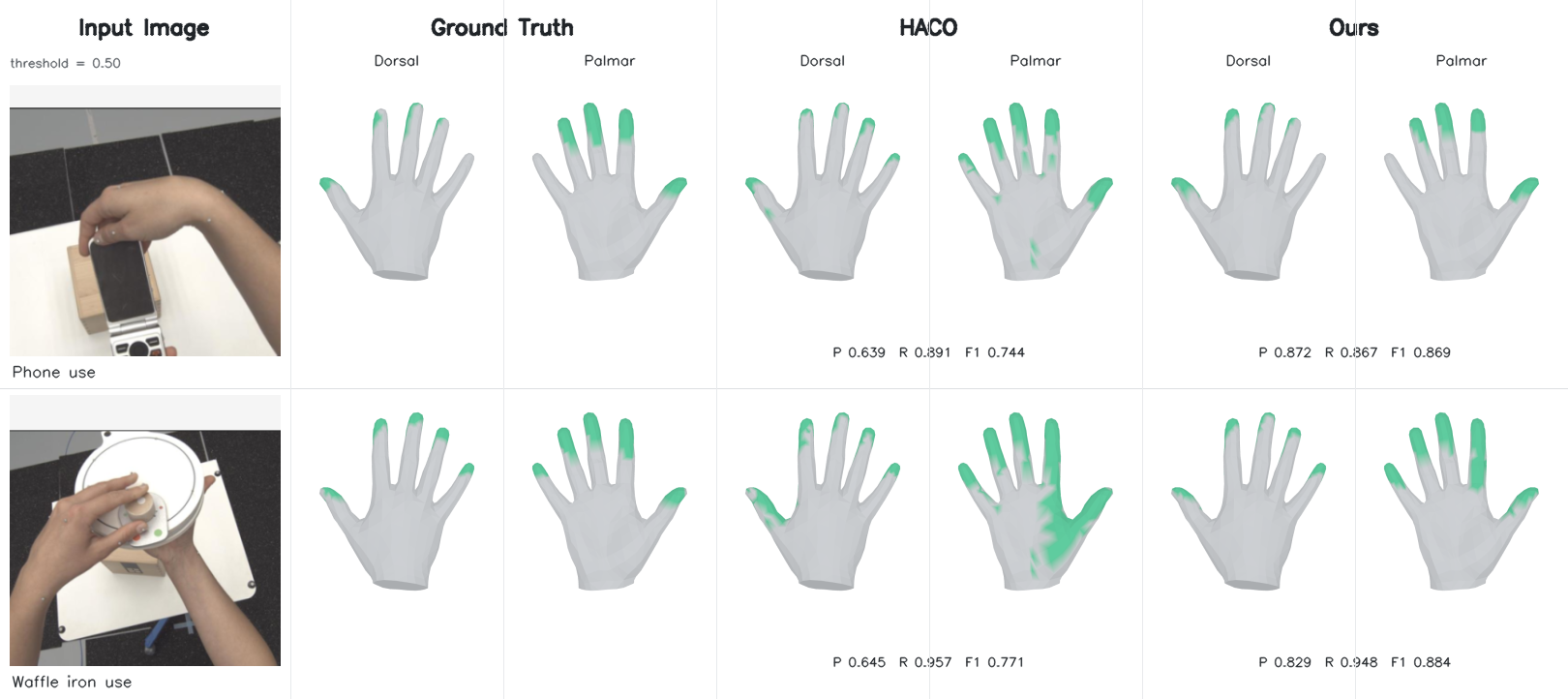}
\caption{Contact visualization on ARCTIC: phone use (frame 332 of
\texttt{s05/phone\_use\_04}) and waffle-iron use (frame 308 of
\texttt{s05/waffleiron\_use\_01}). Both methods use exported GT hand boxes
and contact threshold 0.5. Green marks predicted/reference contact on
canonical hand surfaces, viewed from dorsal and palmar sides; these panels
visualize vertex labels, not recovered hand pose. Compared with the baseline, our method produces cleaner contact regions with fewer spurious predictions. }
\label{fig:app_contact}

\end{figure*}

\begin{figure}[h]
\centering
\includegraphics[width=\textwidth]{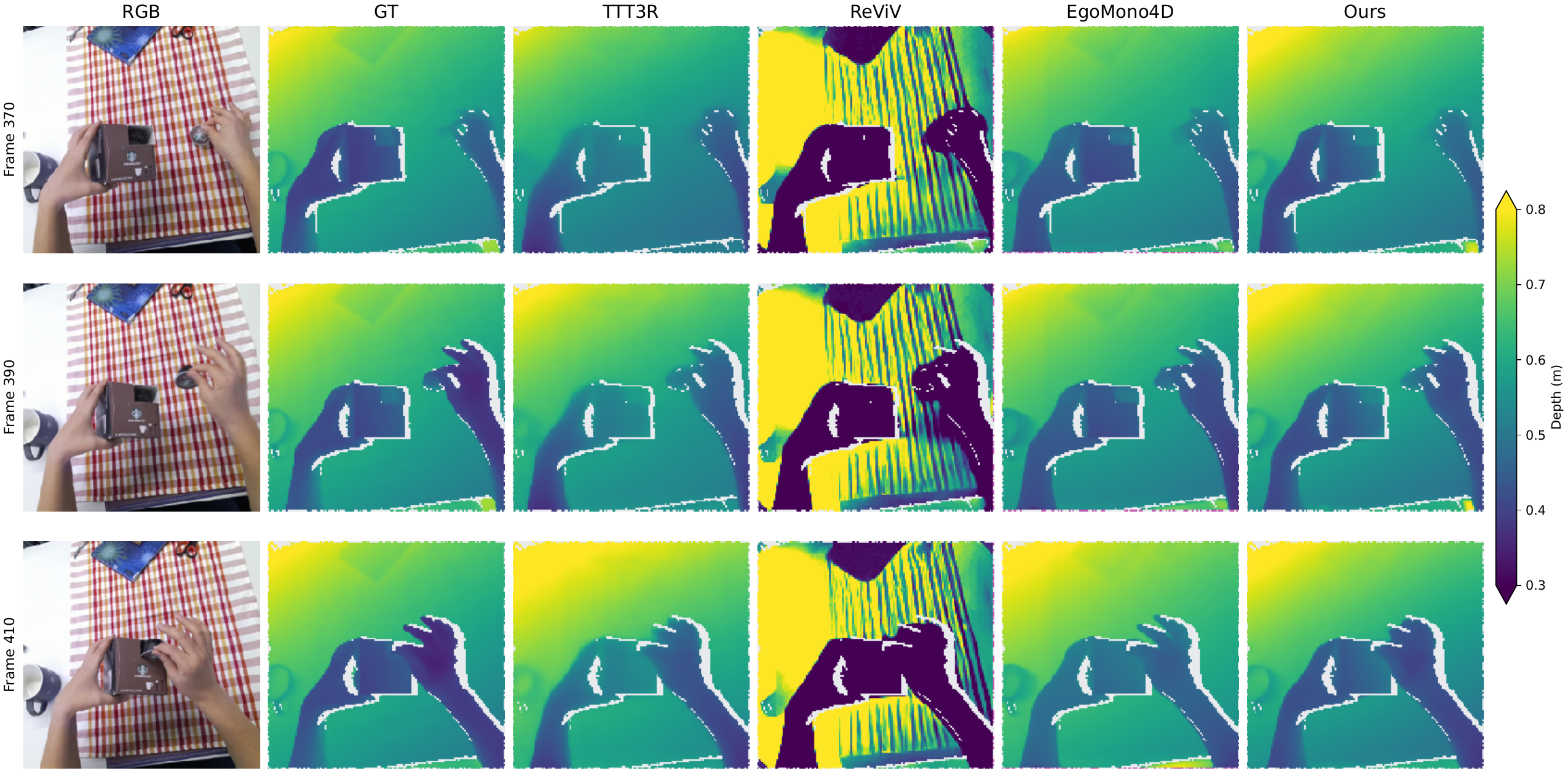}
\caption{H2O depth examples from \texttt{subject4\_ego/h1/0}, frames 370,
390, and 410 of a 60-frame clip. Each method uses one median depth scale
for the whole clip, with the same reference camera rays and common field
of view. Colors share a fixed depth range; unsupported reference regions
are masked. }
\label{fig:app_hot3d}
\end{figure}

Figs.~\ref{fig:app_contact}--\ref{fig:app_hot3d} show some additional qualitative results of contact prediction and depth.

\end{document}